\documentclass[runningheads]{llncs}
\usepackage{graphicx}
\usepackage{amsmath,amssymb} 
\usepackage{color}
\usepackage{subfigure}\usepackage[width=122mm,left=12mm,paperwidth=146mm,height=193mm,top=12mm,paperheight=217mm]{geometry}
\begin{document}
\pagestyle{headings}
\mainmatter
\title{Brain-Inspired Deep Networks for Image Aesthetics Assessment} 

\titlerunning{Brain-Inspired Deep Networks for Image Aesthetics Assessment}

\authorrunning{Zhangyang Wang et.al.}

\author{Zhangyang Wang, Shiyu Chang, Florin Dolcos, Diane Beck, Ding Liu, and Thomas Huang}
\institute{Beckman Institute, University of Illinois at Urbana-Champaign, Urbana, IL}

\maketitle

\begin{abstract}
Image aesthetics assessment has been challenging due to its subjective nature. Inspired by the scientific advances in the human visual perception and neuroaesthetics, we design Brain-Inspired Deep Networks (BDN) for this task. BDN first learns attributes through the parallel supervised pathways, on a variety of selected feature dimensions. A high-level synthesis network is trained to associate and transform those attributes into the overall aesthetics rating. We then extend BDN to predicting the distribution of human ratings, since aesthetics ratings are often subjective. Another highlight is our first-of-its-kind study of label-preserving transformations in the context of aesthetics assessment, which leads to an effective data augmentation approach. Experimental results on the AVA dataset show that our biological inspired and task-specific BDN model gains significantly performance improvement, compared to other state-of-the-art models with the same or higher parameter capacity.
\end{abstract}

\section{Introduction}


Automated assessment or rating of pictorial aesthetics has many applications, such as in an image retrieval system or a picture editing software \cite{cheng2010learning}. Compared to many typical machine vision problems, the aesthetics assessment is even more challenging, due to the highly subjective nature of aesthetics, and the seemingly inherent semantic gap between low-level computable features and high-level human-oriented semantics. Though aesthetics influences many human judgments, our understanding of what makes an image aesthetically pleasing is still limited. Contrary to semantics, an aesthetics response is usually very subjective and difficult to gauge even among human beings.

Existing research has predominantly focused on constructing hand-crafted features that are empirically related to aesthetics. Those features are designed under the guidance of photography and psychological rules, such as rule-of-thirds composition, depth of field (DOF), and colorfulness \cite{ECCV06}, \cite{ke2006design}. With the images being represented by these hand-crafted features, aesthetic classification or regression models can be trained on datasets consisting of images associated with human aesthetic ratings. However, the effectiveness of hand-crafted features is only empirical, due to the vagueness of certain photographic or psychologic rules. Recently, Lu et.al. \cite{rapid} proposed the \textit{Rating Pictorial Aesthetics using Deep Learning} (\textbf{RAPID}) model, with impressive accuracies on the \textit{Aesthetic Visual Analysis} (\textbf{AVA}) dataset \cite{AVA}. However, they have not yet studied more precise predictions, such as finer-grain ratings or rating distributions \cite{distribution}.


 \begin{figure*}[htbp]
\centering
\begin{minipage}{0.99\textwidth}
\centering{
\includegraphics[width=\textwidth]{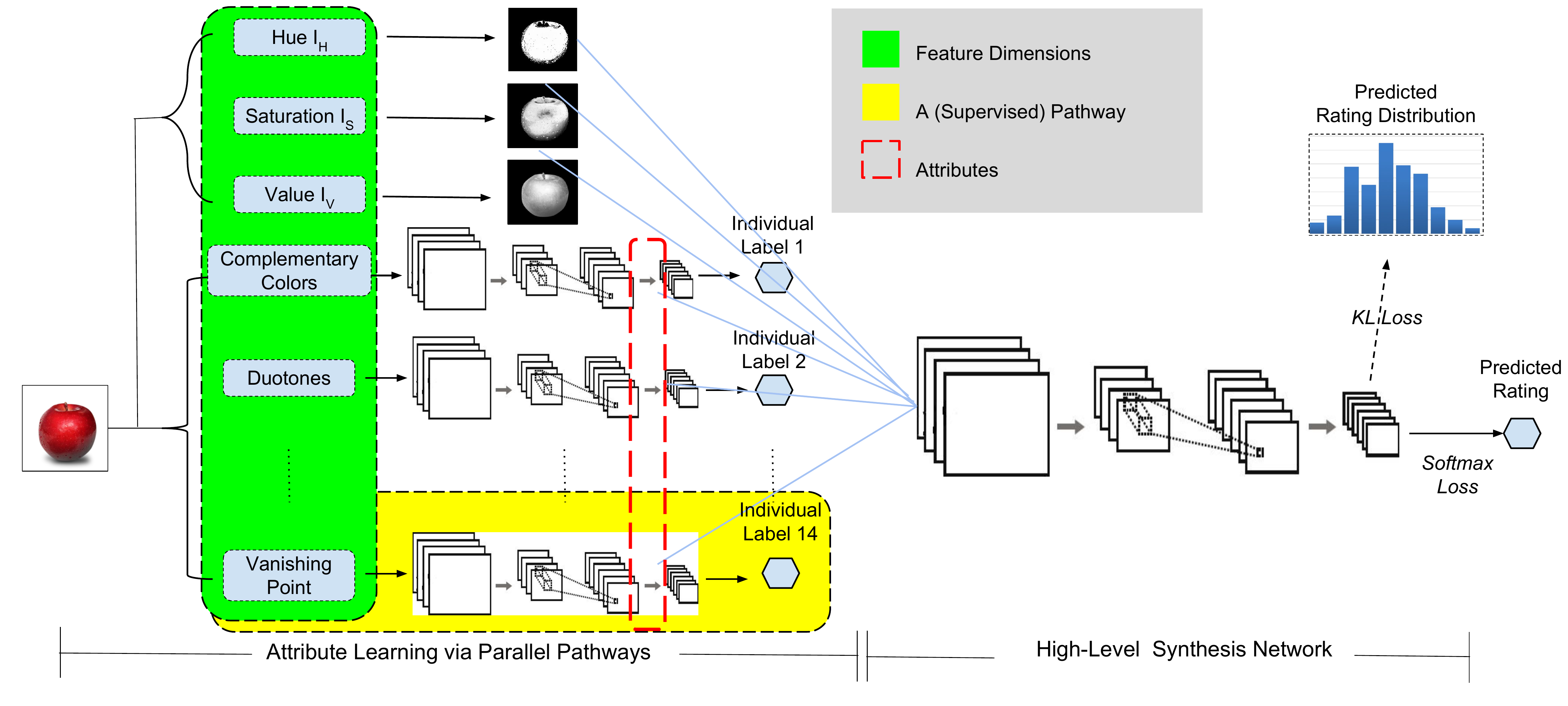}
}\end{minipage}
\caption{The Brain-Inspired Deep Networks (BDN) architecture. The input image is first processed by parallel pathways, each of which learns an attribute along a selected feature dimension independently. Except for the first three simplest features (\textit{hue, saturation, value}), all parallel pathways take the form of fully-convolutional networks, supervised by individual labels; their hidden layer activations are utilized as learned attributes. We then associate those \textit{``pre-trained''} pathways with the high-level synthesis network, and \textit{jointly tune} the entire network to predict the overall aesthetics ratings. In addition to the binary rating prediction, we also extend BDN to predicting the rating distribution, by introducing a Kullback-Leibler (KL)-divergence based loss of the high-level synthesis network.}
\label{DBN}
\end{figure*}

Furthermore, the study of the cognitive and neural underpinnings of aesthetic appreciation by means of neuroimaging techniques yields some promise for understanding human aesthetics \cite{cela2011neural}. Although the results of these studies have been somewhat divergent, a hierarchical set of core mechanisms involved in aesthetic preference have been identified \cite{chatterjee2011neuroaesthetics}. Whereas deep learning is well known to be analogous to brain mechanisms \cite{bengio2009learning}, there is hardly any work providing the synergy between the neuroaesthetics and the advances of learning-based aesthetics assessment models.

In this work, we develop a novel deep-learning based image aesthetics assessment model, called \textbf{Brain-Inspired Deep Networks (BDN)}. BDN clearly distinguishes itself from prior models, for its unique architecture inspired the Chatterjee's visual neuroscience model \cite{prospects}. We introduce the specific architecture of \textit{parallel supervised pathways}, to learn multiple attributes on a variety of selected feature dimensions. Those attributes are then associated and transformed into the overall aesthetic rating, by a \textit{high-level synthesis network}. We extend BDN to predicting the distribution of human ratings, since aesthetics ratings often vary somewhat from observer to observer. Our technical contribution also includes the study of label-preserving transformations in the context of aesthetics assessment, which facilitates data augmentation. We examine the BDN model on the large-scale AVA dataset \cite{AVA}, for both binary rating and rating distribution prediction tasks, and confirms its superiority over a few competitive methods with the same or larger amounts of parameters.

While the neuroscience principles were also considered for traditional aesthetics assessment tasks, BDN makes innovative and meaningful progresses to develop a much more sophisticated and brain-type model, in two ways. First, a deep model by itself, BDN processes the input information in a multiphase hierarchy, which emulates the underlying complex neural mechanisms of human perception. It is more effective and ``biologically plausible'', compared to most standard aesthetics models with hand-crafted features and linear classifiers. Second, among a few existing deep aesthetics assessment models (e.g, RAPID), BDN is the first to introduce the design of independent feature dimensions as parallel pathways, followed by fusing the prediction score. In sum, BDN exploits the neuroaesthetic wisdom (parallel feature extraction, multi-stage prediction, etc.), a part of which were previously utilized only in an oversimplified way, and further integrates such prior wisdom with the power of deep models.


\subsection{Related Work}

Datta et.al. \cite{ECCV06} first casted the image aesthetics assessment problem as a classification or regression problem. A given image is mapped to an aesthetic rating, which is usually collected from multiple subject raters. The rating is normally quantized with discrete values. The earliest work \cite{ECCV06}, \cite{ke2006design} extracted various handcrafted features, including low-level image statistics such as distributions of edges and color histograms, and high-level photographic rules such as the rule of thirds. A part of subsequent efforts, such as \cite{MM10}, \cite{CVPR11}, \cite{ICCV11}, focus on improving the quality of those features. Generic image features \cite{general}, such as SIFT and Fisher Vector \cite{lowe2004distinctive}, have also been applied to predict aesthetics. However, empirical features cannot accurately and exhaustively represent the aesthetic properties.





The human brain transforms and synthesizes a torrent of complex and ambiguous sensory information into coherent thought and decisions. Most aesthetic assessment methods adopt simple linear classifiers to categorize the input features, which is obviously oversimplified. Deep networks \cite{bengio2009learning} attempt to emulate the underlying complex neural mechanisms of human perception, and display the ability to describe image content from the primitive level (low-level features) to the abstract level (high-level features). They are composed of multiple non-linear transformations to yield more abstract and descriptive embedding representations. The RAPID model \cite{rapid} is among the first to apply deep convolutional neural networks (CNN) \cite{imagenet} to the aesthetics rating prediction, where the features are automatically learned. They further improved the model by exploring style annotations \cite{AVA} associated with images. In fact, even the hidden activations from a generic CNN proved to work reasonably well for aesthetics features \cite{ustc}.


Most current work treat aesthetics assessment as a conventional classification problem: the user ratings of each photo are transformed into a ordinal scalar rating (by averaging, etc.), which is taken as the label of the photo. For example, RAPID \cite{rapid} simply divided all samples as aesthetic or unaesthetic, and trained a binary classification model. However, it is common for different users to rate visual subjects inconsistently or even oppositely due to the subjective problem nature \cite{ke2006design}. Since human aesthetic assessment depends on multiple dimensions such as composition, colorfulness, or even emotion \cite{Jiebo}, it is difficult for individuals to reliably convert their experiences to a single rating, resulting in noisy estimates of real aesthetic responses. In \cite{distribution}, Wu et.al. first proposed to represent each photo's rating as a distribution vector over basic ratings, constituting a structural regression problem. Gao et.al. \cite{gao2015multiple} formulated the aesthetic assessment as a multi-label task, where multiple aesthetic attributes were predicted jointly via bayesian networks.

\subsection{Datasets}

Large and reliable datasets, consisting of images and corresponding human ratings, are the essential foundation for the development of machine assessment models. Several Web photo resources have taken advantage of crowdsourcing contributions, such as Flickr and DPChallenge.com \cite{AVA}. The AVA dataset is a large-scale collection of images and meta-data derived from DPChallenge.com. It contains over 250,000 images with aesthetic ratings from 1 to 10, and a 14,079 subset with binary style labels (e.g., rule of thirds, motion blur, and complementary colors), making automatic feature learning using deep learning approaches possible. In this paper, we focus on AVA as our research subject.

\section{Biological Inspirations}

\subsection{Summary of Scientistic Advances}

Recent advances in neuroaesthetics imply that the human perception of aesthetics is a very complicated and systematic process. Multiple parallel processing strategies, involving over a dozen retinal ganglion cell types, can be found in the retina. Each ganglion cell type tiles the retina to focus on one specific kind of feature, and provide a complete representation across the entire visual field \cite{nassi2009parallel}. Retinal ganglion cells project in parallel from the retina, through the lateral geniculate nucleus of the thalamus to the primary visual cortex. Primary visual cortex receives parallel inputs from the thalamus and uses modularity, defined spatially and by cell-type specific connectivity, to recombine these inputs into new parallel outputs. Beyond primary visual cortex, separate but interacting dorsal and ventral streams perform distinct computations on similar visual information to support distinct behavioural goals \cite{functional}. The integration of visual information is then achieved progressively. Independent groups of cells with different functions are brought into temporary association, by a so-called ``binding'' mechanism \cite{cela2011neural}, for the final decision-making.

From the retina to the prefrontal cortex, the human visual processing system will first conduct a very rapid holistic image analysis \cite{treisman1980feature}, \cite{itti1998model}, \cite{tsotsos2011computational}. The divergence comes at a later stage, in how the low-level visual features are further processed through parallel pathways \cite{field2007information} before being utilized. The pathway can be characterized by a hierarchical architecture, in which neurons in higher areas code for progressively more complex representations by pooling information from lower areas. For example, there is evidence \cite{rousselet2004parallel} that neurons in V1 code for relatively simple features such as local contours and colors, whereas neurons in TE fire in response to more abstractive features, that encode the scene's gist and/or saliency information and act as a holistic signature of the input.

\noindent \textbf{Key Notations:} For the consistency of terms, we use \textit{feature dimension} to denote a prominent visual property, that is relevant to aesthetics judgement. We define an \textit{attribute} as the learned abstracted, holistic feature representation over a specific feature dimension. We define a \textit{pathway} as the processing mechanism from a raw visual input to an attribute.

\subsection{Principled Design Insights}

The computational model of deep learning is known to be (loosely) tied to a class of theories of brain development \cite{elman1998rethinking}. For example, the design of CNNs follows the discovery of general human vision mechanisms \cite{functional}, indicating the usefulness of ideas borrowed from neurobiological processes. On the other hand, all current ``deep'' models remain extremely simple compared to the vastness and complexity of biological information processing. It is demonstrated that a single neuron is probably more complex than an entire CNN \cite{stoodley2009functional}, not to mention our lack of knowledge in the cells' electrochemical properties and inter-neuron interactions. We argue that it is neither impractical nor necessary, for model to exactly reproduce the full perception process in the human brain, take the typical example of man being able to fly without the complexity and fluidity of flapping wings. 

The main insights for BDN were gained from the classical and important \textbf{Chatterjee's visual neuroscience model} \cite{prospects}. It models the cognitive and affective processes involved in visual aesthetic preference, providing a means to organize the results obtained in the 2004-2006 neuroimaging studies, within a series of information-processing phases. The Chatterjee's model concludes the following simplified, but important insights, that inspire our model:
\begin{itemize}
\item The human brain works as a multi-leveled system.
\item For the visual sensory input, a variety of relevant feature dimensions are first targeted.
\item A set of parallel pathways abstract the visual input. Each pathway processes the input into an attribute on a specific feature dimension.
\item The high-level association and synthesis transforms all attributes into an aesthetics decision.
\end{itemize}
Step 2 and 3 are derived from the many recent advances \cite{nassi2009parallel} showing that aesthetics judgments evidently involve multiple pathways, which could connect from related perception tasks \cite{cela2011neural}, \cite{chatterjee2011neuroaesthetics}. Previously, many feature dimensions, such as color, shape, and composition, have already been discovered to be crucial for aesthetics. A bold yet rational assumption is thus made by us, that the attribute learning for aesthetics tasks could be decomposed onto those pre-known feature dimensions and processed in parallel.

\section{Brain-Inspired Deep Networks}

The architecture of Brain-Inspired Deep Networks (BDN) is depicted in Fig. \ref{DBN}. The whole training process is divided in two stages, based on the above insights. In brief, we first learn attributes through parallel (supervised) pathways, over the selected feature dimensions. We then combine those \textit{``pre-trained''} pathways with the high-level synthesis network, and \textit{jointly tune} the entire network to predict the overall aesthetics ratings. The testing process is completely feed-forward and end-to-end.

 \begin{table}[t]
\small
\begin{center}
\caption{The 14 style attribute annotations in the AVA dataset}
\label{style}
\begin{tabular}{|c|c||c|c|}
\hline
Style & Number & Style & Number \\
\hline
\hline
Complementary Colors & 949 & Duotones & 1, 301\\
\hline
High Dynamic Range & 396 &  Image Grain & 840 \\
\hline
Light on White & 1,199 &  Long Exposure & 845 \\
\hline
Macro & 1,698 & Motion Blur & 609 \\
\hline
Negative Image & 959 &  Rule of Thirds & 1,031\\
\hline
Shallow DOF & 710 & Silhouettes & 1,389 \\
\hline
Soft Focus & 1,479 & Vanishing Point & 674\\
\hline
\end{tabular}
\end{center}
\end{table}

\subsection{Attribute Learning via Parallel Pathways}

\subsubsection{Selecting Feature Dimensions}

We first select feature dimensions that are discovered to be highly related to aesthetics assessment. Despite the lack of firm rules, certain visual features are believed to please humans more than others \cite{ECCV06}. We take advantage of those photographically or psychologically inspired features as priors, and force BDN to ``focus'' on them.

The previous work, e.g., \cite{ECCV06}. has identified a set of aesthetically discriminative features. It suggested that the light exposure, saturation and hue play indispensable roles. We assume the RGB data of each image is converted to HSV color space, as $I_H$, $I_S$, and $I_V$, where each of them has the same size as the original image\footnote{In our experiments, we downsample $I_H$, $I_S$, and $I_V$ to 1/4 of their original size, to improve the training efficiency. It turns out that the model performance is hardly affected, which is understandable since the human perceptions of those features are insensitive to scale changes.}. Furthermore, many photographic style features influence human's aesthetic judgements. \cite{ECCV06} proposed six sets of photographic styles, including the rule of thirds composition, textures, shapes, and shallow depth-of-field (DOF). The AVA dataset comes with a more enriched variety of \textit{style annotations}, as listed in Table \ref{style}, which are leveraged by us. \footnote{The 14 photographic styles are chosen specifically on the AVA datasets. We do not think they represent all aesthetics-related visual information, and  plan to have more photographic styles annotated.}

%
%
%
%


\subsubsection{Parallel Supervised Pathways}

Among the 17 feature dimensions, the simplest three, $I_H$, $I_S$, and $I_V$ are immediately obtained from the input. However, the remaining 14 style feature dimensions are not qualitatively well-defined; their attributes are not straightforward to be extracted. 

\begin{figure}[tbpht]
\centering
\begin{minipage}{0.80\textwidth}
\centering {
\includegraphics[width=\textwidth]{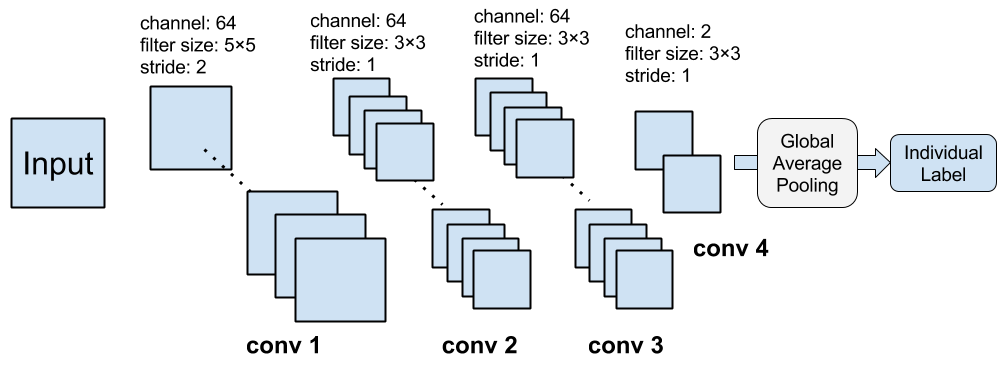}
}\end{minipage}
\caption{The architecture of one supervised pathway, in the form of FCNN. A 2-way softmax classifier is employed after the global averaging pooling, to predict the individual label (0 or 1).}
\label{fcnn}
\end{figure}
For each style category as a feature dimension, we create binary \textit{individual labels}, by labelling images with the style annotation as ``1'' and otherwise ``0'', which follows many previous work \cite{AVA}, \cite{CVPR11}. We design a special architecture, called \textit{parallel supervised pathways}. Each pathway is modeled with a \textit{fully convolutional neural network} (FCNN), as in Fig. \ref{fcnn}. It takes an image as the input, and outputs image's individual label along this feature dimension. All pathways are learned in parallel without intervening with each other. The choice of FCNN is motivated by the spatial locality-preserving property of human brain's low-level visual perception \cite{functional}.



%


For each feature dimension, the number of labeled samples is limited, as shown in Table \ref{style}. Therefore, we pre-train the first two layers in Fig. \ref{fcnn}, using all images from the AVA dataset, in a unsupervised way. We construct a 4-layer Stacked Convolutional Auto Encoder (SCAE): its first 2 layers follows the same topology as the conv1 and conv2 layers, and the last 2 layers are mirror-symmetrical deconvolutional layers \cite{zeiler2011adaptive}. After SCAE is trained, the first two layers are applied to initialize the conv1 and conv2 layers for all 14 FCNN pathways. The strategy is based on the common belief that the lower layers of CNNs learn general-purpose features, such as edges and contours, which could be adapted for extensive high-level tasks \cite{decaf}.

After the initialization of the first two layers, for each pathway, we concatenate the conv3 and conv4 layers, and further conduct supervised training using individual labels. The conv4 layer always has the same channel number with the corresponding style classes (here the channel number is 2 for all, since we only have binary labels for each class). It is followed by the global average pooling \cite{nin} step, to be correlated with the binary labels. Eventually, the conv4 layer as well as the classifier are discarded, and the conv1-conv3 layers of 14 pathways are passed to the next stage. We treat the conv3 layer activations of each pathway as learned attributes \cite{decaf}.



\subsection{Training The High-Level Synthesis Network}

Finally, we simulates brain's high-level association and synthesis, using a larger FCNN. Its architecture resembles Fig. \ref{fcnn}, except that the first three convolutional layers each have 128 channels instead of 64. The high-level synthesis network takes the attributes from all parallel pathways as inputs, and outputs the overall aesthetics rating. The entire BDN is then tuned from end to end.

\section{Predicting The Distribution Representation}

Most existing studies \cite{ECCV06} apply a scalar value to represent the predicted aesthetics quality, which appears insufficient to capture the true subjective nature. For example, two images with the equal mean score could have very different deviations among raters. Typically, an image with a large rating variance is more likely to be edgy or subject to interpretation. \cite{rapid} assigned images with binary aesthetics labels,  i.e., high quality and low quality, by thresholding their mean ratings, which provided less informative supervision due to the large intra-class variation. \cite{distribution} suggested to represent the ratings as a distribution on pre-defined ordinal basic ratings. However, such a structural label could be very noisy, due to the coarse grid of basic ratings, the limited sample size (number of ratings) per image, and the lack of shifting robustness of their $L_2$-based loss.

The previous study of the AVA datasets \cite{AVA} reveals two important facts:
\begin{itemize}
\item For all images, the standard deviation of an image's ratings is a function of its mean rating. Especially,  images with ``moderate'' ratings  tend to have a lower variance than images with ``extreme'' ratings. It inspires us that the estimations of mean ratings and standard deviations may be jointly performed, which can potentially mutually reinforce each other.
\item For each image, the distribution of its ratings from different raters is largely Gaussian. According to \cite{AVA}, Gaussian functions perform adequately good approximations to fit the rating distributions of 99.77\% AVA images. Besides, those non-Gaussian distributions tend to be highly-skewed, occurring at the low and high extremes of the rating scale, where their mean ratings could be predicted with higher confidences.
\end{itemize}
To this end, we propose to explicitly model the rating distribution for each image as Gaussian, and jointly predict its mean and standard deviation. Assuming the underlying distribution $N_1(\mu_1, \sigma_1)$ and the predicted distribution $N_2(\mu_2, \sigma_2)$, their difference is calculated by the Kullback-Leibler (KL) divergence \cite{bishop2006pattern}:
\begin{equation}
\begin{array}{l}\label{KL}
KL(N_1, N_2) = \log\frac{\sigma_2}{\sigma_1} + \frac{\sigma_1^2 + (\mu_1 - \mu_2)^2}{2\mu_2^2} - \frac{1}{2}
\end{array}
\end{equation}
$N_1$ is calculated by fitting the rating histogram (over the 10 discrete ratings) of each image, with a Gaussian model. It is treated as  the ``ground truth'' here. $KL(N_1, N_2) $ = 0 if and only if the two distributions are exactly the same, and increases while $N_2$ diverges from $N_1$. 

When training BDN to predict rating distributions, we replace the default softmax loss with the loss function (\ref{KL}), which corresponds to the KL-loss branch (the dash) in Fig. \ref{DBN}. The outputs of the global average pooling from the high-level synthesis network remains to be a vector $\in R^{2 \times 1}$. But different from the binary prediction task where the output denotes a Bernoulli distribution over [0, 1] labels, the two elements in the output here denote the predicted mean and variance, respectively. They could thus be arbitrary real values falling within the rating scale. 


\begin{figure*}[tbp]
\centering
\begin{minipage}{0.45\textwidth}
\centering \subfigure[] {
\includegraphics[width=\textwidth]{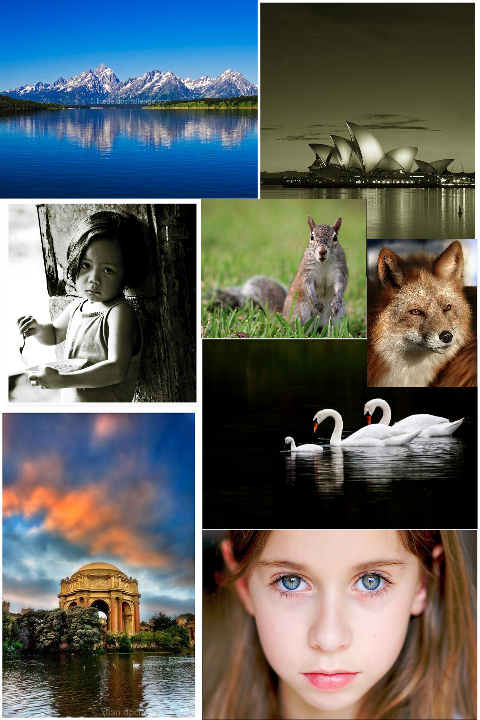}
}\end{minipage}
\begin{minipage}{0.45\textwidth}
\centering \subfigure[] {
\includegraphics[width=\textwidth]{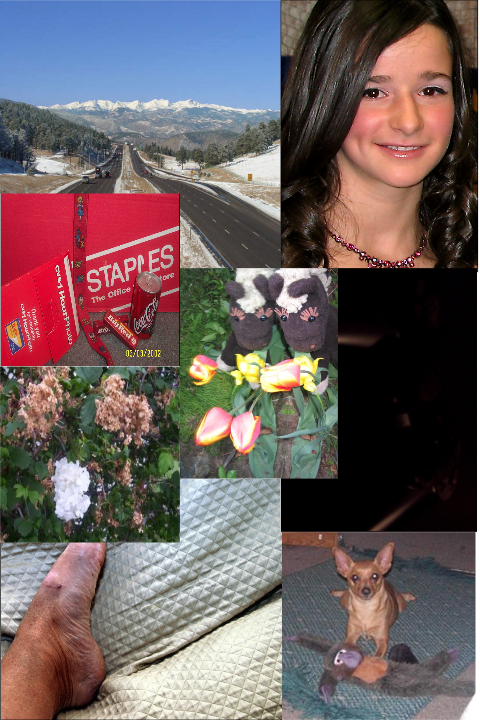}
}\end{minipage}
\caption{BDN classification examples: (a) high-quality; (b) low-quality ($\delta$ = 0).}
\label{image}
\end{figure*}

\section{Study Label-Preserving Transformations}

 When training deep networks, the most common approach to reduce overfitting is to artificially enlarge the dataset using label-preserving transformations \cite{bishop2006pattern}. In \cite{imagenet}, image translations and horizontal reflections are generated, while the intensities of the RGB channels are altered, both of which apparently will not change the object class labels. Other alternatives, such as random noise, rotations, warping and scaling, are also widely adopted by the latest deep-learning based object recognition methods. However, there has been little work on identifying label-preserving transformations for image aesthetics assessment, e.g., those that will not significantly alter the human aesthetics judgements, considering the rating-based labels are very subjective. In \cite{rapid}, motivated by their need to create fixed-size inputs, the authors created randomly-cropped local regions from training images, which was empirically treated as data augmentation.


 \begin{table}[bp]
  \scriptsize
\begin{center}
\caption{The subjective evaluation survey on the aesthetics influences of various transformations ($s$ denotes a random number)}
\label{trans}
\begin{tabular}{|c|c||c|}
\hline
Transformation & Description & LP factor   \\
\hline
Reflection & Flipping the image horizontally   &0.99 \\
\hline
Random scaling & Scale the image proportionally by $s \in$ [0.9, 1.1]  & 0.94 \\
\hline
Small noise & Add a Gaussian noise $\in$ $N(0, 5)$ & 0.87 \\
\hline
Large noise & Add a Gaussian noise $\in$ $N(0, 30)$ & 0.63 \\
\hline
Alter RGB & Perturbed the intensities of the RGB channels \cite{imagenet} &  0.10 \\
\hline
Rotation & Randomly-parameterized affine transformation & 0.26 \\
\hline
Squeezing & Change the aspect ratio by $s \in$ [0.8, 1.2] & 0.55 \\
\hline
\end{tabular}
\end{center}
\end{table}

We make the first exploration to identify whether a certain transformation will preserve the binary aesthetics rating, i.e., high quality versus low quality, by conducting a \textbf{subjective evaluation survey} among over 50 participants. We select 20 high-quality ($\delta$ = 1) images from the AVA dataset (since low-quality images are unlikely to become more aesthetically pleasing after some simple/random transformations). Each image is processed by all different kinds of transformations in Table \ref{trans}. For each time, a participant is shown with a set of image pairs originated from the same image, but processed with different transformations (including the groundtruth). For each pair, the participant needs to decide which one is better in terms of aesthetics quality. The image pairs are drawn randomly, and the image winning this pairwise comparison will be compared again in the next round, until the best one is selected. 

We fit a Bradley-Terry \cite{bradley1952rank} model to estimate the subjective scores for each method so that they can be ranked. With groundtruth set as score 1, each transformation will receive a score between [0, 1]. We define the score as the \textit{label-preserving} (\textbf{LP}) factor of a transformation; a larger LP factor denotes a smaller impact on image aesthetics. According to Table \ref{trans}, \textit{reflection} and \textit{random scaling} receive high LR factors; the small noise seems to marginally affect the aesthetics feelings, while all the remaining will significantly degrade human aesthetics perceptions. We therefore adopt reflection, random scaling, and small noise as our default data augmentation approaches, unless otherwise specified.

\section{Experiment}

\subsection{Settings}

We implement our models based on the cuda-convnet package \cite{imagenet}. The ReLU nonlinearity as well as dropout is applied. The batch size is fixed as 128. Since BDN is fully convolutional, there is no need to normalize the input size. Experiments run on a workstation with 12 Intel Xeon 2.67GHz CPUs and 1 GTX680 GPU. Training one pathway takes roughly 4-5 hours. The fine-tuning of the entire BDN model typically takes about one day. 

For binary prediction, we follow RAPID \cite{rapid} to quantize images' mean ratings into binary values. Images with mean ratings smaller than $5 - \delta$ are labeled as ``low-quality'', while those with mean ratings larger than $5 + \delta$ are referred to as ``high-quality''. For the distribution prediction, we do not quantize the ratings.

The adjustment of learning rates in such a hierarchical model calls for special attentions. We first train the 14 parallel pathways, with the identical learning rates: $\eta$ = 0.05 for unsupervised pre-training and 0.01 for supervised tuning, both of which are not annealed throughout training. We then train the high-level synthesis network on top of them and fine-tune the entire BDN. For the pathway part, its learning rate $\eta'$ starts from 0.001; for the high-level part, the learning rate $\rho$ starts from 0.01. When the training curve reaches a plateau, we first try dividing $\rho$ by 10; and further try dividing $\rho$ by 10 if the training/validation error still does not decrease.

\noindent \textbf{Static Regularization versus Joint Tuning} The RAPID model \cite{rapid} also extracted attributes along different columns (pathways) and combine them. The pre-trained style classifier was then ``frozen'' and acted as a static network regularization. Out of curiosity, we also tried to fix our parallel pathways while training the high-level synthesis network, e.g., $\eta'$ = 0. The resulting performance was verified to be inferior to that of joint tuning the entire BDN. 

\subsection{Binary Rating Prediction}

We compare BDN with the state-of-the-art RAPID model for binary aesthetics rating prediction.
Benefiting from our fully-convolutional architecture, the BDN model has a much lower parameter capacity than RAPID that relies on fully-connected layers. In addition, we compare the proposed model to three baseline networks, all with exactly the same parameter capacity as BDN:
\begin{itemize}
\item \textbf{Baseline fully-convolutional network (BFCN)} first binds the conv1 -- conv3 layers of 14 pathways horizontally, constituting a three-layer fully convolutional network, each layer owning 64 $\times$ 14 = 896 filter channels. The attribute learning part is trained in a unsupervised way, and then concatenated with the high-level synthesis network, to be jointly supervised-tuned. BFCN does not utilize style annotations.
\item \textbf{BDN without parallel pathways (BDN-WP)} utilize style annotations in an entangled fashion. Its only difference with BFCN lies in that, the training of the attribute learning part is supervised by a composite label $\in R^{28 \times 1}$, which binds 14 individual labels altogether.
\item \textbf{BDN without data augmentations (BDN-WA)} denotes BDN without the three data augmentations applied (reflection, scaling, and small noise).
\end{itemize}
We train the above five models for binary rating predictions, with both $\delta$ = 0 and $\delta$ = 1. The overall accuracies are compared in Table \ref{overall}. \footnote{The accuracies of RAPID are from the RDCNN results in Table 3 \cite{rapid}} It appears that BFCN performs significantly worse than others, due to the absence of the style attribute information. While RAPID, BDN-WP and BDN all utilize style annotations as the supervision, BDN outperforms the other two in both cases with remarkable margins. By comparing BDN-WP with BDN, we observe that the biologically-inspired parallel pathway architecture in BDN facilitates the learning. Such a specific architecture avoids overly large all-in-one models (such as BDN-WP), but instead have more effective, dedicated sub-models. In BDN, style annotations serve as powerful priors, to enforce BDN to focus on extracting features that are highly correlated to aesthetics judgements. The BDN is jointly tuned from end to end, which is different from RAPID whose style column only acts as a static regularization. We also notice a gain of nearly 3\% of BDN over BDN-WA, which verifies the effectiveness of our proposed augmentation approaches.

In \cite{AVA}, a linear classifier was trained on fisher vector signatures computed from color and SIFT descriptors. Under the same aesthetic quality categorization setting, the baselines reported by \cite{AVA} were 66.7\% when $\sigma$ = 0, and 67.0\% when $\sigma$ = 1, falling far behind both BDN and RAPID.

\begin{table}[t]
\begin{center}
\caption{The accuracy comparison of different methods for binary rating prediction.}
\label{overall}
\begin{tabular}{|c|c|c|c|c|c|}
\hline
& RAPID & BFCN & BDN-WP & BDN-WA & BDN   \\
\hline
$\delta$ = 0 & 74.46\%  & 70.20\% & 73.54\% & 74.03\% & 76.80\% \\
\hline
$\delta$ = 1 & 73.70\% & 68.10\% &  72.23\% & 73.72\% & 76.04\% \\
\hline
\end{tabular}
\end{center}
\end{table}

\begin{figure}[bpht]
\centering
\begin{minipage}{0.50\textwidth}
\centering \subfigure[] {
\includegraphics[width=\textwidth]{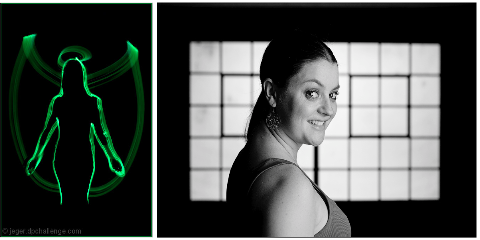}
}\end{minipage}
\begin{minipage}{0.40\textwidth}
\centering \subfigure[]  {
\includegraphics[width=\textwidth]{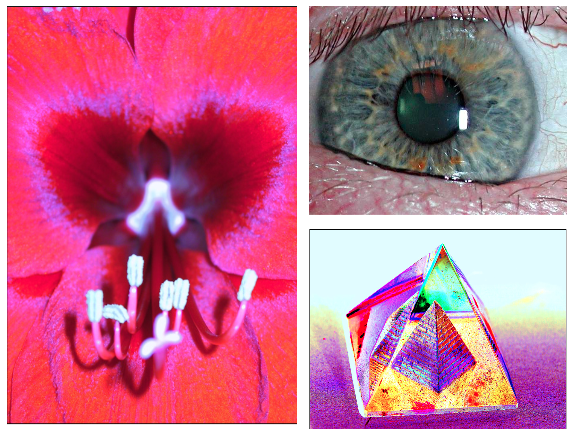}
}\end{minipage}
\caption{How contexts and emotions could alter the aesthetics judgment. (a) Incorrectly classified examples ($\delta$ = 0) due to semantic contents; (b) High-variance examples (correctly predicted by BDN), which have nonconventional styles or subjects.}
\label{failure}
\end{figure}

To qualitatively analyze the results, we display eight images correctly classified by BDN to be high-quality when $\delta$ = 0, in Fig. \ref{image} (a), and eight correctly classified low-quality images in in Fig. \ref{image} (b). The images ranked high in terms of aesthetics typically present salient foreground objects, low depth of field, proper composition, and color harmony. In contrast, low-quality images are at least defected in one aspect. For example, the top left image has no focused foreground object, while the bottom right one suffers from a messy layout. For the top right ``girl'' portrait in Fig \ref{image} (b), we investigated its original comments on DPChallenge.com, and found that people rated it low because of the noticeable detail loss caused by noise reduction post-processing, as well as the unnatural ``plastic-like'' lights on her hair.

Even more interestingly, Fig. \ref{failure} (a) lists two \textbf{failure} examples of BDN. The left image in Fig. \ref{failure} (a) depicts a waving glowstick captured by time-lapse photography. The image itself has no appealing composition or colors, and is thus identified by BDN to be low-quality. However, the DPChallenge raters/commenters were amazed by the angel shape and rated it very favorably due to the creative idea. The right image, in contrast, is a high-quality portrait, on which DBN confidently agrees.  However, it was associated with the ``Rectangular'' challenge topic on DPChallenge, and was rated low because this targeted theme was overshadowed by the woman. The failure examples manifest the huge subjectivity and sensitivity of human aesthetics judgement.


\subsection{Rating Distribution Prediction}
\begin{table}[t]
\begin{center}
\caption{The average KL divergence comparison for rating distribution prediction.}
\label{KL}
\begin{tabular}{|c|c|c|}
\hline
 BDN & BDN-soft-D & BDN-KL-D   \\
\hline
0.1743  & 0.2338 & 0.2052 \\
\hline
\end{tabular}
\end{center}
\end{table}

To our best knowledge, among all state-of-the-art models working on latest large-scale datasets, BDN is the only one accounting for rating distribution prediction. We use the binary prediction BDN as the initialization, and re-train only the high-level synthesis network with the loss defined in Eqn. (\ref{KL}). We then compare the predicted distributions with the groundtruth of the AVA testing set. We also include two more BDN variants as baselines in this task:
\begin{itemize}
\item \textbf{BDN with the softmax loss for rating distribution vectors (BDN-soft-D)} makes the only architecture change by modifying the global average pooling of the high-level network to be 10-channel. Its output is compared to the raw rating distribution under the conventional softmax loss (i.e., cross entropy).
\item \textbf{BDN with the KL loss for rating distribution vectors (BDN-KL-D)} replaces the softmax loss in BDN-soft-D, with the general KL loss (i.e., relative entropy) \cite{bishop2006pattern}. It remains to work with the raw rating distribution.
\end{itemize}
As compared in Table \ref{KL}, KL-based loss function tends to perform better than the softmax function for this specific task. It is important to notice that BDN further reduces the KL divergence compared to BDN-KL-D. While the raw ratings can be noisy due to both the coarse rating grid and the limited rating number, we are able to obtain a more robust estimation of the underlying rating distribution, with the aid of the strong Gaussian prior from the AVA study \cite{AVA}.

Very notably, we observe that for more than 96\% of the AVA testing images, the differences between their groundtruth mean values and estimates by BDN are less than 1. We further binarize the estimated and groundtruth mean values, to re-evaluate the results in the context of binary rating prediction. The overall accuracies are improved to 78.08\% ($\delta$ = 0), and 77.27\% ($\delta$ = 1). It verifies the benefits to jointly predict the means and standard deviations, built upon the AVA observation that they are correlated.



Fig. \ref{failure} (b) visualizes images that are correctly predicted by BDN to have large variances. It is intuitive that images with a high variance seem more likely to be edgy or subject to interpretation. Taking the top right image for example, the comments it received indicate that while many voters found the photo striking (e.g. ``nice macro''``good idea''), others found it rude (e.g. ``it frightens me''``too close for comfort'').

%

\section{Discussions}

There have been efforts continued to explore distinct aspects of the neural underpinnings of aesthetic appreciation, such as recognition and familiarity \cite{fairhall2008neural}, bottom-up versus top-down pathways \cite{cupchik2009viewing}, and the influence of expertise \cite{calvo2010experts}. A few of them could also be corresponded to the computational process in BDN. For example, the bottom-up/top-down pathways \cite{cupchik2009viewing} reminds the feed-forward/back-propogration processes in training deep networks. 

There is certainly much room to strengthen the synergy between neuroaesthestics and computaitonal models. The findings in \cite{jacobsen2010beauty} indicated that aesthetic judgements partially overlap with the evaluative judgements on social and moral cues, which is also implied by our examples in Fig. \ref{failure}. Our immediate next work is to take them into account. 

\section{Conclusion}

In this paper, we get inspired by the knowledge abstracted from the human visual perception and neuroaesthetics, and formulate the Brain-Inspired Deep Networks (BDN). The biological inspired, task-specific architecture of BDN leads to superior performances, compared to other state-of-the-art models with the same or higher parameter capacity. Since it has been observed in Fig. \ref{failure} that emotions and contexts could alter the aesthetics judgment, we plan to take the two factors into account for a more comprehensive framework.


\bibliographystyle{splncs}
\bibliography{egbib}
\end{document}